\documentclass[a4paper,conference]{IEEEtran}

\ifCLASSINFOpdf
  \usepackage[pdftex]{graphicx}
  
\else

\fi

\usepackage{times}
\usepackage{soul}
\usepackage{url}
\usepackage[utf8]{inputenc}
\usepackage[small]{caption}
\usepackage{graphicx}
\usepackage{amsmath}
\usepackage{amsthm}
\usepackage{booktabs}
\usepackage{algorithm}
\usepackage{algorithmic}

\usepackage{amssymb} 
\usepackage{float} 
\usepackage{stfloats} 
\usepackage{hyperref}
\hypersetup{hypertex=true,
            colorlinks=true,
            linkcolor=black,
            anchorcolor=black,
            citecolor=blue}

\title{Interactive Image Inpainting Using \\Semantic Guidance}

\graphicspath{{Figures/}}

\author{
\IEEEauthorblockN{Wangbo Yu,
Jinhao Du, 
Ruixin Liu, 
Yixuan Li
and
Yuesheng Zhu\thanks{* Yuesheng Zhu is the corresponding author.}}
\IEEEauthorblockA{Communication and Information Security Lab\\
Shenzhen Graduate School, Peking University, China\\
wbyu@stu.pku.edu.cn, jinhaodu@stu.pku.edu.cn, anne\underline{\space}xin@pku.edu.cn, liyixuan99@stu.pku.edu.cn, zhuys@pku.edu.cn}}

\begin{document}

\maketitle

\begin{abstract}
  Image inpainting approaches have achieved significant progress with the help of deep neural networks. However, existing approaches mainly focus on leveraging the priori distribution learned by neural networks to produce a single inpainting result or further yielding multiple solutions, where the controllability is not well studied. This paper develops a novel image inpainting approach that enables users to customize the inpainting result by their own preference or memory. Specifically, our approach is composed of two stages that utilize the prior of neural network and user's guidance to jointly inpaint corrupted images. In the first stage, an autoencoder based on a novel external spatial attention mechanism is deployed to produce reconstructed features of the corrupted image and a coarse inpainting result that provides semantic mask as the medium for user interaction. In the second stage, a semantic decoder that takes the reconstructed features as prior is adopted to synthesize a fine inpainting result guided by user's customized semantic mask, so that the final inpainting result will share the same content with user's guidance while the textures and colors reconstructed in the first stage are preserved. Extensive experiments demonstrate the superiority of our approach in terms of inpainting quality and controllability.
\end{abstract}

\section{Introduction}
Image inpainting refers to the task of filling in missing regions of a corrupted image with visually realistic and contextually coherent contents. Early works~\cite{PatchMatch,pic5,pic7,pic13} mainly focus on leveraging the contextual information of the damaged image to repair, where although the contextual continuity is ensured, the synthesized content often lacks meaningful semantic information. With the rapid progress of CNNs and GANs~\cite{GAN}, several deep learning-based approaches~\cite{CE,GntIpt,SF,RFR} are proposed, which could produce semantically meaningful results utilizing the learned priors from training data. These approaches target for single image generation and are not capable of generating various results. More recently,~\cite{UCTGAN,VQVAE,PDGAN} have devoted to design one-to-many image mapping models to realize multi-solution inpainting. Although the diversity of generated results is realized, they ignore a key factor in image inpainting, that is controllability. Depart from above approaches, we argue that image inpainting tasks should be highly related to users, who ought to be enabled to subjectively specify the inpainting results based on their own preference or memory, instead of passively accepting a single inpainting result or selecting a barely satisfied one from multiple results produced by models. Therefore, in this paper, we focus on designing an image inpainting framework that enable users to customize the inpainting result. 

Compared with single-solution image inpainting and multi-solution image inpainting, such controllable image inpainting task still lacks detailed discussing. Although several two-stage inpainting approaches~\cite{EC,SPG} have the potential to be promoted to this task, both of them are designed to firstly convert damaged image into completed guidance information (edge or semantic mask) and then use the guidance to inpaint, where the two stages are totally seperate and too much useful information like colors and textures are discarded in the first stage, so that their controllability and inpainting results are far from satisfying. Image editing approches like ~\cite{faceshop,SCFEGAN,GC} may also be applied to this task. However, these approaches are bound to take guidance information as model input in the beginning to controll the editing result, so that users have to be highly aware of what they expect to present in the missing region and draw the corresponding guidance starting from scratch, which will place high demand on users' drawing skill especially when inpainting complex images. 

In contrast to above approaches, we propose a novel interactive image inpainting approach comprised of two closely connected stages that utilize the prior of neural network and user's guidance to jointly inpaint damaged images. In the first stage, we design an autoencoder to produce a coarse inpainting result which will then be passed into a well-trained semantic segmentation network to provide semantic guidance for user interaction, lest users should draw the guidance starting from scratch. In the second stage, a decoder with several spatially adaptive normalization (SPADE)~\cite{spade} layers is adopted, which will accept the reconstructed features from the first stage as input, and generate a fine result according to the semantic guidance customized by users, so that the final inpainting result will have the same content with users' customized semantic guidance, while the colors and textures reconstructed in the first stage are preserved. Our approach faces two challenges: firstly, the quality of semantic guidance served as interactive medium is highly dependent on the autoencoder, so that the coarse inpainting result of which should be as accurate as possible; secondly, the downsample process of the autoencoder will lead to loss of the contextual information and thus affect the performance of the second stage which is conditioned on the encoded features. To tackle these challenges, We further apply an external spatial attention module (dubbed ESPA) in the autoencoder, it models long range dependency between the context of the damaged image and its features, realizing spatial information propagation via a lightweight external attention mechanism operated on the spatial dimension, and thus promoting the quality of both the semantic guidance and the features fed to the second stage.  

Our contributions are summarized as follows:
\begin{itemize}
\item We propose a novel interactive image inpainting approach that utilizes the prior of models and users' guidance to jointly inpaint the damaged image.
\item We propose an external spatial attention module (ESPA) that integrates context and encoded features of the damaged image together and realize spatial information propagation using a lightweight external attention mechanism to enhance the inpainting quality.
\item Experiments on challenging datasets including FFHQ~\cite{ffhq}, Celeba-HQ~\cite{celebahq}, and Outdoorscenes~\cite{outdoor} demonstrate that our approach not only achieves superior performance in inpainting quality, but also possesses strong capability in controllability. 
\end{itemize}

Our source code and demo will be made publicly available at \url{https://github.com/Mark-Yu1/Interactive-Image-Inpainting}.

\section{Related Works}
\subsection{Guidance Information Based Image Inpainting}
In recent image inpainting approaches, several kinds of guidance information were introduced to provide structural information for the inpainting network and enhance the quality of the inpainting result. To name a few, the single-channel edge map which uses binary values to characterize the contour of an image was introduced by~\cite{EC}. ~\cite{SF} proposed to use three-channel smooth image to supply more robust structural information rather than using edge as guidance, and thus achieved better performance.~\cite{SPG} proposed to utilize the multi-channel semantic mask which contains the class information of every pixel to eliminate the blurry boundaries between different objects in the inpainting area. The above approaches are all designed to translate the damaged image into completed guidance information firstly and use the completed guidance to inpaint, where the two stages are totally separate and too much useful information like colors and textures are discarded in the first stage. Recently,~\cite{SGE} and~\cite{SGE+} further applied the semantic guidance to reconstruct correct structures for complex inpainting area with mixed scenes. Although both of them utilized one-stage model to avoid loss of useful information, the controllability is lost.

\subsection{Attention Mechanism in Image Inpainting} 
Attention mechanisms, especially spatial attention mechanisms were broadly adopted in recent inpainting approaches to model the correlation between the missing region and contextual information of the corrupted image. To name a few,~\cite{GntIpt} proposed contextual attention which processes the similarity between background patches and the coarse inpainting result of missing region, and utilizes the most matching patches to further refine the coarse inpainting result.~\cite{csa} utilized a semantic coherent attention layer which not only propagates information from context to missing region, but also models the relationship between patches inside the inpainting area. In~\cite{UCTGAN}, cross semantic attention was adopted to compute pixel-wise attention between the instance image and corrupted image at feature level, and perform weighted reconstruction on the deep feature of corrupted image using the computed attention. The above attention mechanisms face one glaring shortcoming that the computational complexity of which are too high due to the massive matrix multiplication operations. To this end, we propose a novel external spatial attention module that models long range dependency between the contextual information and reconstructed features of the corrupted image using two external parameter matrices, which are shared across samples and bringing linear complexity for attention mechanism. External attention was firstly introduced in~\cite{EANet} and originally applied at channel dimension. We further apply it at spatial dimension and reform the structure of external parameter matrices as well as the calculation process to make it suitable for inpainting task, which is detailed in Section 3.

\subsection{Semantic Guidance in Low-level Vision Tasks} 
Recent researches have demonstrated that the multi-channel semantic masks produced by semantic segmentation models could help to guide low-level vision tasks. For example, In image generation tasks,~\cite{p2p},~\cite{crn},~\cite{sgsg},~\cite{p2phd} use semantic masks as prior condition for image translation. What's more, ~\cite{spade} proposes spatially adaptive de-normalization (SPADE) to guide the semantic image generation process to avoid the vanishment of semantic information in forward propagating process, it could synthesize vivid natural scenes guided by corresponding semantic mask. In other low-level vision tasks such as image super resolution~\cite{SFTSR}, image denoising~\cite{smdns}, Neural style transfer~\cite{a2r} and image manipulation~\cite{mskgan},~\cite{mskpe}, the semantic guidance has also been adopted to promote the quality of synthesized images. In our inpainting approach, we adopt the semantic guidance for two reasons: firstly, the semantic guidance could serve as geometric constraint to ensure the inpainting result has clear boundary. Secondly, as the synthetic result will share the same geometry with the semantic guidance, we utilize the semantic mask as the medium for user interaction, where users could customize the semantic mask and get the corresponding inpainting result.

\begin{figure*}[!htb] 
  \centering
  \includegraphics[width=2.05\columnwidth]{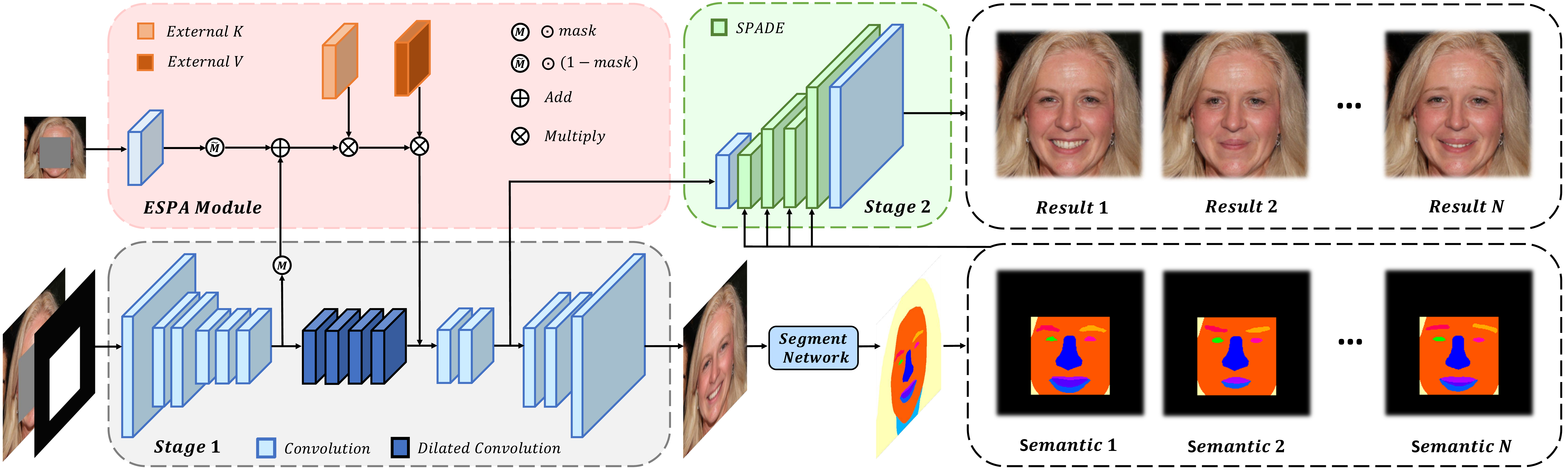}
  \caption{Overall architecture of the proposed framework. In stage 1, an autoencoder based on ESPA module is deployed to produce reconstructed features of the corrupted image and a coarse inpainting result that provides semantic mask as the medium for user interaction. In stage 2, a semantic decoder that takes the reconstructed features as prior is adopted to synthesize a fine inpainting result guided by user's customized semantic mask, so that the final inpainting result will share the same content with user's guidance, while the textures and colors reconstructed in stage 1 are preserved.}
  \label{fig:architecture}
  \end{figure*}

\section{Approach}
Figure \ref{fig:architecture} depicts the overall architecture of our framework, which consists of three parts: an autoencoder based on external spatial attention (ESPA), a semantic segmentation network and a decoder based on spatially adaptive normalization (SPADE)~\cite{spade}. The inpainting process can be divided into two stages: In the first stage, the ESPA autoencoder produces a coarse inpainting result which will then be passed into the semantic segmentation network to provide semantic mask for user interaction; In the second stage, the semantic decoder accepts the encoded features from the ESPA autoencoder as input, and will synthesize a particular fine inpainting result guided by the semantic mask customized by users. 

\subsection{ESPA Autoencoder}
\subsubsection{Network Architecture} 
The ESPA autoencoder consists of a convolutional encoder, a convolutional decoder and a two-branch bottleneck layer, where one branch contains several dilated convolution layers, and another branch is mainly based on the ESPA module. 

Let $E$ be the autoencoder, $\boldsymbol{I} \in \mathbb{R}^{3 \times H \times W}$ denote the corrupted image and $\boldsymbol{M} \in \mathbb{R}^{1 \times H \times W}$ denote the binary mask used for marking the damaged region (1 for damaged pixels and 0 for the rest). We use masked image $\boldsymbol{I}_{in}=\boldsymbol{I} \odot (1-\boldsymbol{M})$ ($\odot$ denotes the elementwise multiplication) and its corresponding mask $\boldsymbol{M}$ as the input of $E$, the coarse output of $E$ is formulated as
\begin{equation}
 \boldsymbol{I}_{c}=E(\boldsymbol{I}_{in}, \boldsymbol{M}).
 \end{equation} 
\subsubsection{External Spatial Attetion Module} 
The ESPA module serves as another bottleneck branch paralleling with the dilated convolution layers, it integrates context and encoded features of the damaged image together and computing long range dependency between them. Different from attention mechanisms used in~\cite{GntIpt,csa,UCTGAN} that compute attention weights internally, the ESPA module utilizes two external learnable key and value matrices that multiply with the input query along dimension $H$ and dimension $W$ respectively to realize spatial information propagation. The external key matrix and value matrix share the same structure that is simply built upon two linear layers along with an activation function, they are independent of the input query feature and shared across samples, improving the generalization capability and bring linear complexity for the attention mechanism.

Specifically, ESPA module is applied on the bottleneck feature $\boldsymbol{F}_{in} \in \mathbb{R}^{c \times \frac{H}{4} \times \frac{W}{4}}$. In order to match the size of $\boldsymbol{F}_{in}$, We downsample the context image $\boldsymbol{I}_{in}$ and then apply a $1\times 1$ convolution to reshape it into $\boldsymbol{I}_{sub} \in \mathbb{R}^{c \times \frac{H}{4} \times \frac{W}{4}}$. The final input query of the external attention is:
\begin{equation}
  Q = \boldsymbol{I}_{sub} \odot (1-\boldsymbol{M}_{sub}) + \boldsymbol{F}_{in} \odot \boldsymbol{M}_{sub}.
\end{equation}
Here, $\boldsymbol{M}_{sub} \in \mathbb{R}^{1 \times \frac{H}{4} \times \frac{W}{4}}$ denotes the corresponding downsampled mask. The computing process of external attention can be formulated as: 
\begin{equation}
  ESPA(Q) = (Q^{T} \cdot \tilde{K})^{T} \cdot \tilde{V},
\end{equation}
where $\tilde{K}$ and $\tilde{V}$ represent the external key and value matrices respectively.
\subsubsection{Loss Functions}
In the first stage, we exploit L1 distance as the image reconstruction loss, and further adopt perceptual loss~\cite{perloss} to enhance the similarity between the synthesized image and its ground truth in semantic feature level.

Let $\boldsymbol{I}_{gt}$ be the ground truth image, the reconstruction loss is expressed as:
\begin{equation}
\mathcal{L}_{rec} = \mathbb{E}_{I}[\parallel \boldsymbol{I}_{gt} - \boldsymbol{I}_{c} \parallel_1].
\end{equation}

The perceptual loss is expressed as:
\begin{equation}
  \begin{aligned}
  \mathcal{L}_{per} = & \mathbb{E}_{I}[\sum_{j \in l} \frac{1}{C_jH_jW_j} \parallel \phi^j(\boldsymbol{I}_{gt}) - \phi^j(\boldsymbol{I}_{c}) \parallel^2_2], 
  \end{aligned}
  \end{equation}
where $l$ denotes the selected layers of VGG-19~\cite{vgg}, $\phi^j(\cdot)$ is the $j$-th layer's output feature and $C_j$, $H_j$, $W_j$ denote the number of its channels, height, width respectively. 

The total loss of the first stage can be formulated as:
\begin{equation}
  \begin{aligned}
    \mathcal{L}_{stage1} = \lambda_{rec} \mathcal{L}_{rec} + \lambda_{per} \mathcal{L}_{per},
  \end{aligned}
  \end{equation}
we set $\lambda_{rec} = \lambda_{per} = 1$ in our experiments.
\subsection{Semantic Decoder}
\subsubsection{Network Architecture}
The semantic decoder is built on several convolution layers and spatially adaptive normalization (SPADE)~\cite{spade} layers. It is conditioned on the encoded feature of the ESPA autoencoder, and guided by multi-channel semantic mask produced by the semantic segmentation network. For semantic segmentation networks, we utilize a well-trained DeepLab V2~\cite{deeplab} for natural scene segmentation, and a BiSeNet~\cite{bisenet} for parsing face images. For the sake of interactivity, we binarize the multi-channel probability maps produced by segmentation networks into hard labels, so that users could directly interact with the corresponding pseudo-color image using drawing tools, and the customized result will then be converted to multi-channel semantic mask to guide the semantic decoder. 

Let $G$ be the semantic decoder, denote the modified semantic mask as $\boldsymbol{S}_{m}$. By conditioning on the encoded feature $\boldsymbol{F}_{c}$ of the ESPA autoencoder, the final inpainting result is expressed as:
\begin{equation}
  \boldsymbol{I}_{f}=G(\boldsymbol{F}_{c}, \boldsymbol{S}_{m}).
  \end{equation} 
\subsubsection{Loss Functions}
In the second stage, we use ground truth image $\boldsymbol{I}_{gt}$ and its semantic mask $\boldsymbol{S}_{gt}$ to establish injective relationship between the semantic guidance and the inpainting result.  Besides reconstruction loss $\mathcal{L}_{rec}$ and perceptual loss $\mathcal{L}_{per}$ used in the first stage, we further apply an $70 \times 70$ patchGAN~\cite{p2p} discriminator, denoted as $D$, to provide adversarial loss that enriches detailed textures for the final inpainting result. We use the least-squares GAN~\cite{lsgan} for stable training, the adversarial loss is expressed as:
\begin{equation}
  \begin{aligned}
    \mathcal{L}_{adv} = \mathbb{E}[(D(\boldsymbol{I}_{gt}))^2] + \mathbb{E}[(1 - D(G(\boldsymbol{F}_{c}, \boldsymbol{S}_{gt})))^2].
  \end{aligned}
  \end{equation}
The total loss of the second stage is formulated as:
\begin{equation}
  \begin{aligned}
    \mathcal{L}_{stage2} = \lambda_{rec} \mathcal{L}_{rec} + \lambda_{per} \mathcal{L}_{per} + \lambda_{adv} \mathcal{L}_{adv},
  \end{aligned}
  \end{equation}
where the hyper-parameters $\lambda_{rec}$, $\lambda_{per}$ and $\lambda_{adv}$ are set to $1$, $1$ and $0.01$ respectively. 
\section{Experiments}
\subsection{Experimental Settings}
We conduct experiments on three challenging datasets: FFHQ~\cite{ffhq}, Celeba-HQ~\cite{celebahq}, and Outdoorscenes~\cite{outdoor}. We resize all the images into $256\times 256$ for training and testing to make fair comparison with existing approaches. Both regular masks and irregular masks~\cite{PC} are adopted for training to enable our model to handle different types of damaged images. As mentioned in~\cite{GntIpt} that image inpainting approaches lack proper evaluation metrics since there are multiple solutions for an damaged image. Nevertheless, we still evalute our model on PSNR~\cite{psnr},SSIM~\cite{ssim} and FID~\cite{fid} to make quantitative comparison. We use  128$\times$128 center mask for all quantitative evaluations. Our models are implemented by PyTorch and trained on two NVIDIA 2080Ti GPUs. The training process could be divided into three stages: In the first stage, we train the ESPA autoencoder to get deep features of corrupted image and a coarse inpainting result that provide semantic mask for user interaction. In the second stage, the semantic decoder was trained to utilize the feature prior and semantic guidance to synthesize a specific fine inpainting result.
\begin{figure}[H] 
  \centering
  \includegraphics[width=0.98\columnwidth]{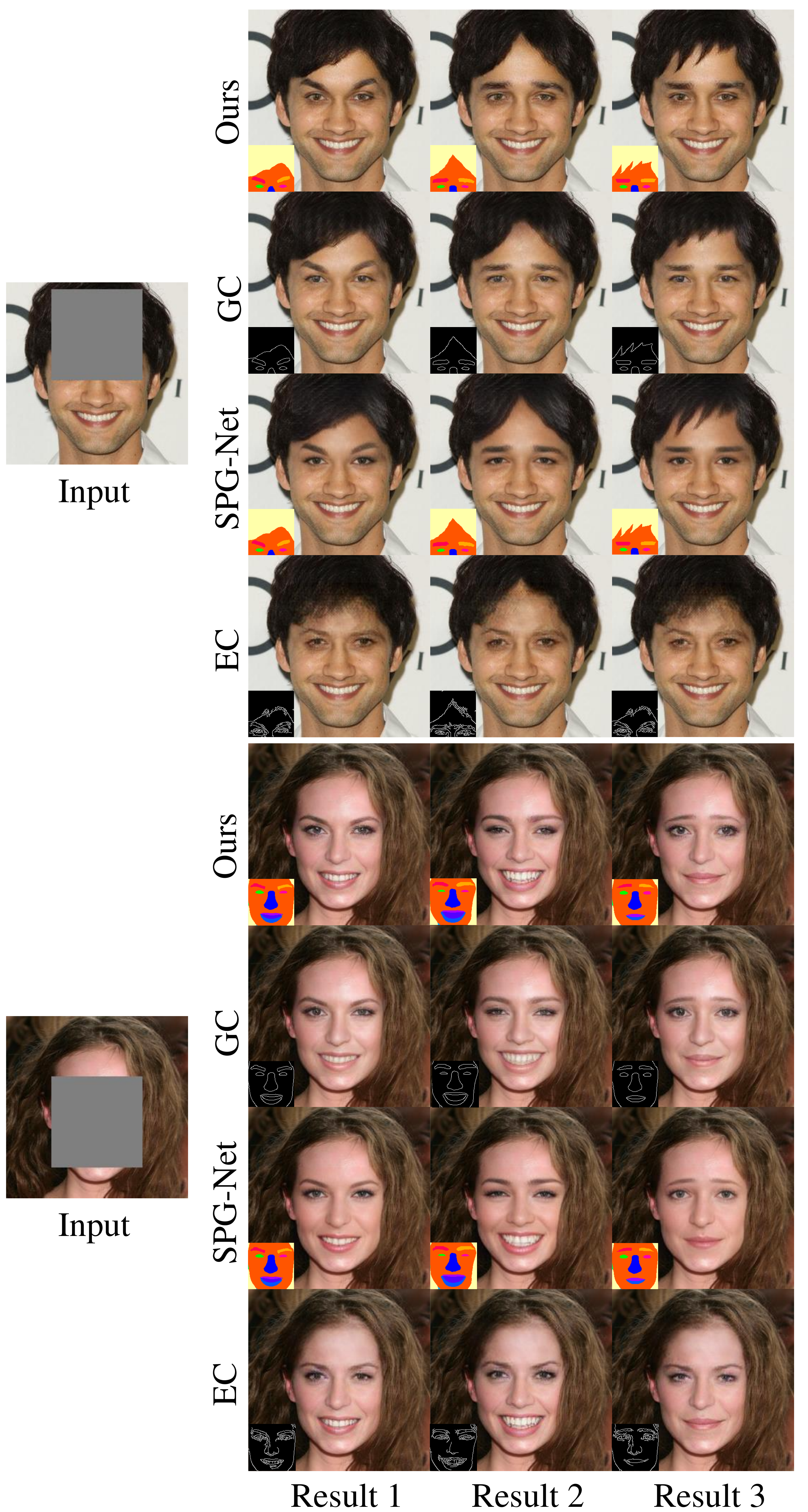}
  \caption{Qualitative comparison with guidance-based approaches. The guidances are shown in the left-bottom corner of every image.}
  \label{fig:cnl}
  \end{figure}
\begin{table}[H]
  \centering
  \scalebox{1.2}{
  \begin{tabular}{c|ccc} 
  \toprule
  Approach & PSNR $\uparrow$ & SSIM $\uparrow$ & FID $\downarrow$  \\
  \hline 
   EC & 25.75 & 0.8793 & 8.04 \\
   SPG-Net & 27.92 & 0.8943 &  6.85  \\
   GC & 26.51 & 0.9048 & 5.22\\
   Ours& \textbf{28.09} & \textbf{0.9232} & \textbf{3.86} \\
  \bottomrule
  \end{tabular}}
  \caption{Quantitative comparison with guidance-based approaches. All the approaches are tested using ground truth guidance.}
  \label{tab:quantcnl}
  \end{table}
Finally, we perform joint training of the two stages to further improve the inpainting quality of our framework. During training, Adam optimizer~\cite{adam} was adopted with detail momentum settings: $\beta_1 = 0.0, \beta_2 = 0.9$. For all the training stages, the batch size was set to 1, the learning rate was set to 0.0002 and fixed until 500,000 iterations, then, it linearly decayed to 0 until 1,000,000 iterations. 
\begin{figure*}[t] 
  \centering
  \includegraphics[width=2.03\columnwidth]{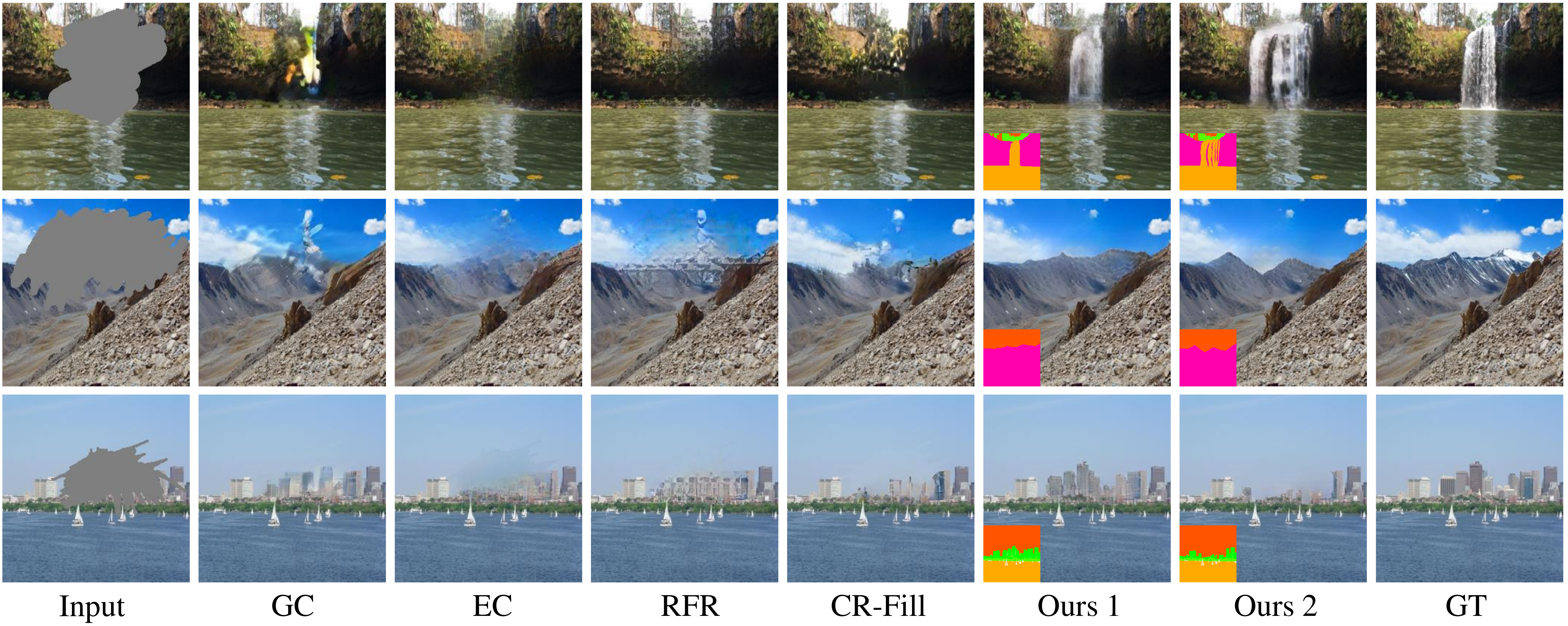}
  \caption{Comparison with state-of-the-arts on Outdoorscenes~\cite{outdoor} dataset. We show two different inpainting results of our approach according to the semantic guidance in the left-bottom corner.}
  \label{fig:sota}
  \end{figure*}
\begin{table}[tb]
    \centering
    \scalebox{1.2}{
  \begin{tabular}{c|ccc} 
    \toprule
    Approach & PSNR $\uparrow$ & SSIM $\uparrow$ & FID $\downarrow$  \\
    \hline 
     GC & 19.06 & 0.73 & 42.43 \\
     EC & 19.32 & 0.75 & \textbf{41.25}  \\
     RFR & 20.83 & 0.75 & 42.08\\
     CR-Fill & 21.36 & 0.78 & 41.90\\
     Ours& \textbf{22.20} & \textbf{0.81} & 41.65\\
    \bottomrule
    \end{tabular}}
    \caption{Quantitative comparison with state-of-the-art approaches.}
    \label{tab:sota}
    \end{table}
\subsection{Evaluation on Controllability}
Firstly, We evaluate the controllability of our approach on Celeba-HQ~\cite{celebahq} dataset. The comparison baselines are three guidance-based approaches: EC~\cite{EC}, an edge-guided two-stage model that completes edge map firstly and uses the completed edge as guidance to inpaint corrupted image secondly; GC~\cite{GC}, also an edge guided two-stage model yet the edge guidance is totally drawn by users and directly served as model input of the first stage, while the second stage could be regarded as a refine process. we re-implement GC because the edge-guided model is unavailable in their official code. SPG-Net~\cite{SPG}, a two-stage model guided by semantic mask, the inpainting process of which follows the same idea with EC, i.e., to complete the guidance firstly and inpaint image secondly. We also re-implement SPG-Net since it is not open source. The subjective results are shown in Figure. \ref{fig:cnl}.

For fair comparison, we directly utilize ground truth guidances derived from three different ground truth images to experiment on the same input corrupted image, the guidances are shown in the left-bottom corner of every inpainting results. Specifically, EC uses edge maps produced by Canny edge detector~\cite{canny} as guidance, which is very messy and thus affecting the inpainting quality and controllability. SPG-Net uses the same semantic guidance with our approach, although it could synthesize correct structures according to the given guidance, it fails to reconstruct detailed textures. The smooth edge map in GC is originally made by detecting landmarks of faces and manually connecting nearby landmarks into lines, which costs a lot of manual labours and the hair region is ignored due to the limitation of the key point detection algorithm utilized in their paper. To solve these problems, in our re-implementation, we apply Canny edge detector on the semantic mask of ground truth images to extract smooth edge map as guidance in GC. Although the textures and colors are reconstructed well in GC, it fails to generate exact structures according to the guidance, since edge maps are single-channel images that are easily ignored and washed off by the network. What's more, GC could not directly offer completed guidance as the medium for user interaction, so that users have to draw the guidance by themselves starting from scratch, placing high demand on their drawing skill. Compared with the above approaches, the reconstructed structure of our approach is exactly the same with semantic guidance, and the reconstructed textures and colors are much closer to the context, since the reconstructed features from the first stage are sufficiently utilized as prior in our approach. Table \ref{tab:quantcnl} shows the quantitative comparison results. We use ground truth guidance for all the tested approaches, in this case, the inpainting task can be regarded as a single-solution problem, so that the quantitative metrics will be suitable for evaluating the inpainting quality. The quantitative results demonstrate that our approach achieves the best inpainting quality.
\begin{figure}[t] 
  \centering
  \includegraphics[width=0.98\columnwidth]{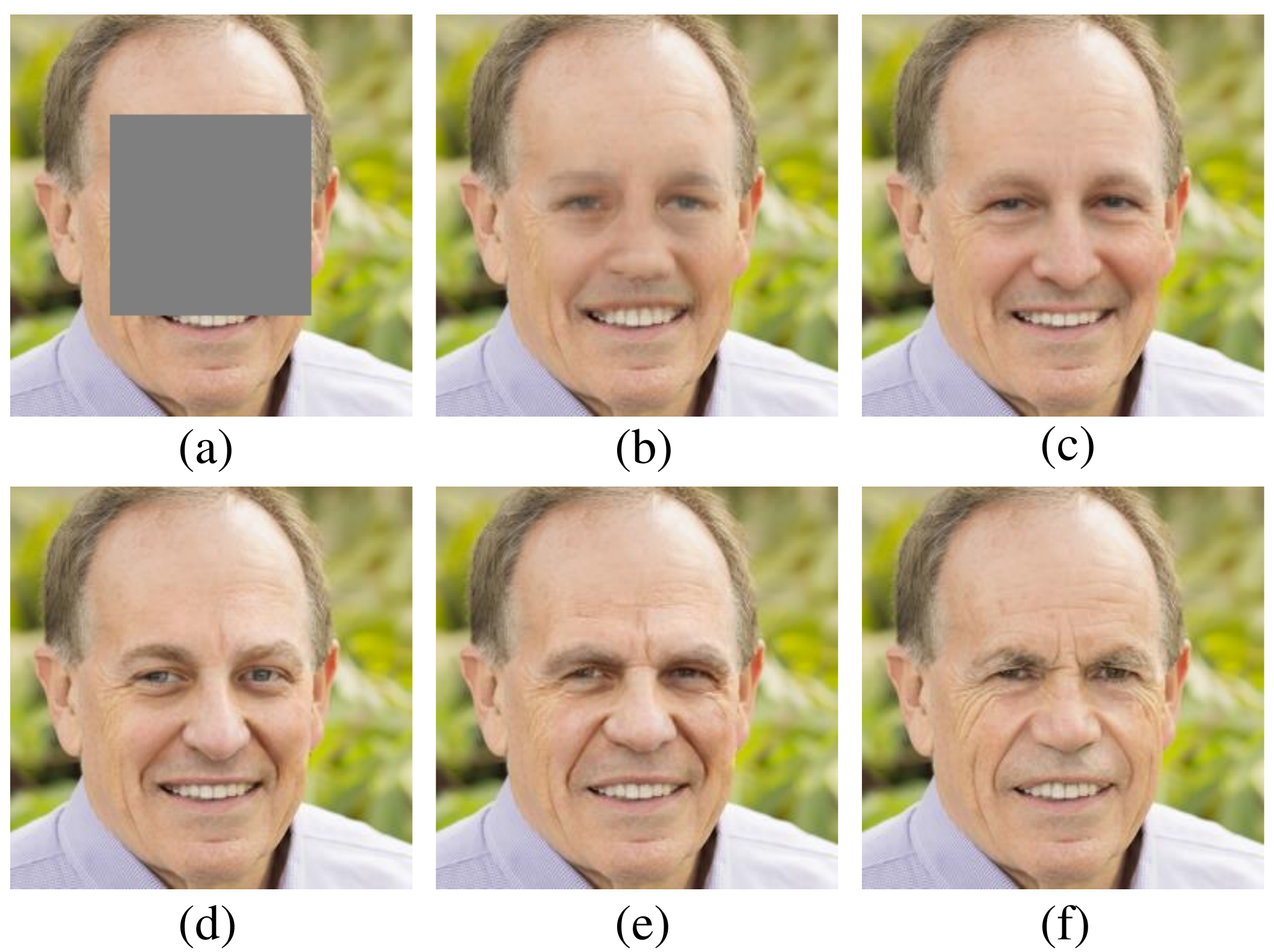}
  \caption{Qualitative results of different settings. (a) Input corrupted image. (b) Result of stage1 without ESPA module. (c) Result of stage1 with ESPA module. (d) Result of stage2 with semantic mask produced by stage1. (e) Result of stage2 with ground truth semantic mask. (f) Ground truth image.}
  \label{fig:ablation}
  \end{figure}
\begin{table}[t]
  \centering
  \scalebox{1.2}{
  \begin{tabular}{c|ccc} 
  \toprule
  Settings & PSNR $\uparrow$ & SSIM $\uparrow$ & FID $\downarrow$  \\
  \hline 
   (b) & 24.97 & 0.8732 & 12.48 \\
   (c) & 26.26 & 0.8980 & 7.58  \\
   (d) & 25.28 & 0.8864 & 8.82 \\
   (e) & \textbf{27.65} & \textbf{0.9148} & \textbf{4.63} \\
  \bottomrule
  \end{tabular}}
  \caption{Quantitative results of different settings. (b) Result of stage1 without ESPA module. (c) Result of stage1 with ESPA module. (d) Result of stage2 with semantic mask produced by stage1. (e) Result of stage2 with ground truth semantic mask.}
  \label{tab:ablation}
  \end{table}
\subsection{Comparison with State-of-the-arts}
We evaluate the performance of our approach on the Outdoorscenes~\cite{outdoor} dataset in comparison with several state-of-the-art approaches, including EC~\cite{EC}, GC~\cite{GC}, RFR ~\cite{RFR} and CR-Fill~\cite{crfill}. The subjective results (Figure. \ref{fig:sota}) demonstrate that our approach not only achieves better inpainting quality beyond other approaches, but also performs well in controllability. Quantitative results are shown in Table \ref{tab:sota}, where our approach achieves the best PSNR~\cite{psnr} and SSIM~\cite{ssim} score, and the second best FID~\cite{fid} score.
\subsection{Ablation Studies}
Finally, we perform ablation studies to evaluate the effectiveness of every module in our model. Specifically, we adopt the FFHQ~\cite{ffhq} dataset which contains 60,000 training face images and 10,000 testing face images to evaluate the effectiveness of the proposed ESPA autoencoder, since the attention mechanism performs better when trained on large datasets~\cite{vit}. We conducted four experiments with different settings: stage1 without ESPA module; stage1 with ESPA module; stage2 with semantic mask produced by stage1; stage2 with ground truth semantic mask. The qualitative results are shown in Figure. \ref{fig:ablation}.

As shown in Figure. \ref{fig:ablation} (b) and (c), given a corrupted image (a), the coarse inpainting result of naive autoencoder without ESPA module exists obvious artifacts and severe blur in the face region, because too much useful contextual information are lost in the downsample process. With the help of external attention module which supply additional contextual information to the deep features of the autoencoder, the ESPA autoencoder could reconstruct desirable inpainting result with clear face region, so that the quality of both the deep features feed to the second stage and the segmentation map served as interactive medium is ensured. With the guidance of semantic mask and deep features reconstructed in the first stage as prior, the semantic decoder of the second stage could refine the coarse inpainting result and synthesize a customized fine result according to the semantic guidance. Typically, if users make no modification on the semantic mask produced by the first stage, the model would directly synthesize a corresponding fine result, as shown in Figure. \ref{fig:ablation} (c) and (d). If users could precisely remember the missing content of the corrupted image and modify the semantic mask close to ground truth semantic mask, then the inpainting result will be extremely close to the ground truth image, as shown in Figure. \ref{fig:ablation} (e) and (f). All in all, users could interact with the semantic mask and modify it by their own preference or memory to get a specific inpainting result.

Table. \ref{tab:ablation} shows the quantitative results of different settings. Interestingly, it is easy to find that the quantitative comparison between (c) and (d) is contrary to the subjective results shown in Figure. \ref{fig:ablation}, where although the fine inpainting result in (d) has clearer textures and less artifacts than the coarse inpainting result in (c), the quantitative results of (d) are worse than that of (c). It further demonstrates that quantitative evaluation metrics are not suitable enough for evaluating ill-posed problems like image inpainting.
\section{Conclusion}
In this paper, we explore a novel two-stage image inpainting framework that utilizes the prior of neural network and user's guidance to jointly inpaint corrupted images. In the first stage, we design an autoencoder with a novel external spatial attention module to produce reconstructed features of the corrupted image and a coarse inpainting result that provides semantic mask as the medium for user interaction. In the second stage, a semantic decoder that takes the reconstructed features as prior is adopted to synthesize a fine inpainting result guided by user's customized semantic mask, so that the final inpainting result will share the same content with user's guidance, while the textures and colors reconstructed in the first stage are preserved. Experiments on various datasets including face images and natural scenes demonstrate the superiority of our approach in terms of inpainting quality and controllability. In future work, we plan to add reference image in our inpainting approach, so that user could not only customize the content of the inpainting result, but also choose their prefer reference image to supply additional color and texture to the inpainting result.

\bibliographystyle{IEEEtran}
\bibliography{./Interactive_Inpaint}

\end{document}